\documentclass[journal]{IEEEtran}

\usepackage{graphicx} 
\usepackage{lineno}
\usepackage[colorlinks,
linkcolor=red,
anchorcolor=red
citecolor=red,
]{hyperref}
\usepackage{bm}
\usepackage{graphics}
\usepackage{subfig}
\usepackage{float}
\usepackage{algorithm}
\usepackage{algorithmicx}
\usepackage{algpseudocode}
\usepackage{setspace}
\usepackage{amsmath,amssymb,amsfonts}
\usepackage{eucal}
\usepackage{enumitem}
\usepackage{array}
\usepackage{epstopdf}
\usepackage{mathtools}
\usepackage{caption}
\usepackage{graphicx, subfig}
\usepackage{chngcntr}
\usepackage{mathrsfs}
\usepackage{makecell,multirow,diagbox}
\usepackage{bm}
\usepackage{colortbl}
\usepackage{xcolor}
\usepackage{ntheorem}
\usepackage{placeins}
\usepackage{tablefootnote}
\usepackage{threeparttable}
\usepackage{makecell}
\usepackage{verbatimbox}
\usepackage{cases}
\usepackage{orcidlink}
\usepackage{booktabs}
\usepackage{mwe}
\usepackage{lipsum}
\usepackage{verbatimbox}
\usepackage{caption}

\usepackage[numbers,sort&compress]{natbib}
\begin{document}
\title{Explicit Change Relation Learning for Change Detection in VHR Remote Sensing Images}
\author{Dalong~Zheng,
	Zebin~Wu,~\IEEEmembership{Senior~Member,~IEEE,}
	Jia~Liu,~\IEEEmembership{Member,~IEEE,}\\
	Chih-Cheng~Hung,~\IEEEmembership{Member,~IEEE,}
	Zhihui~Wei,~\IEEEmembership{Member,~IEEE}
	%\vspace{-0.1cm}
	\thanks{
		Manuscript received DD MM, YY; revised DD MM, YY; accepted DD MM, YY. Date of publication MM DD, YY. 
		This work was supported in part by the National Natural Science Foundations of China under Grant 62071233, Grant 61971223, Grant 62276133, and Grant 61976117; in part by the Jiangsu Provincial Natural Science Foundations of China under Grant BK20211570, Grant BK20180018, and Grant BK20191409; in part by the Fundamental Research Funds for the Central Universities under Grant 30917015104, Grant 30919011103, Grant 30919011402, and Grant 30921011209; in part by the Key Projects of University Natural Science Fund of Jiangsu Province under Grant 19KJA360001; and in part by the Qinglan Project of Jiangsu Universities under Grant D202062032.
		(\textit{Corresponding~author:~Zebin~Wu.})
	}
	\thanks{\IEEEcompsocthanksitem D. Zheng, Z. Wu, J. Liu, and Z. Wei are with the School of Computer Science and Engineering, Nanjing University of Science and Technology (NJUST), Nanjing 210094, China. (E-mail: zhengdl@njust.edu.cn, wuzb@njust.edu.cn, omegaliuj@njust.edu.cn, and gswei@njust.edu.cn.)\protect
		\IEEEcompsocthanksitem C. Hung are with the Center for Machine Vision and Security Research, Kennesaw State University (KSU), USA.
		%Digital Object Identifier XX.XXXX/TGRS.2023.XXXXXXX
}}
\vspace{-0.5cm}	
%Digital Object Identifier XX.XXXX/GRSL.2023.XXXXXXX
\markboth{Journal of \LaTeX\ Class Files,~Vol.~, No.~, MM~YY}%
{Shell \MakeLowercase{\textit{et al.}}: Bare Demo of IEEEtran.cls for IEEE Journals}

\maketitle
\begin{abstract}
Change detection has always been a concerned task in the interpretation of remote sensing images. It is essentially a unique binary classification task with two inputs, and there is a change relationship between these two inputs. 
At present, the mining of change relationship features is usually implicit in the network architectures that contain single-branch or two-branch encoders. 
However, due to the lack of artificial prior design for change relationship features, these networks cannot learn enough change semantic information and lose more accurate change detection performance. So we propose a network architecture NAME for the explicit mining of change relation features.
In our opinion, the change features of change detection should be divided into pre-changed image features, post-changed image features and change relation features. In order to fully mine these three kinds of change features, we propose the triple branch network combining the transformer and convolutional neural network (CNN) to extract and fuse these change features from two perspectives of global information and local information, respectively. In addition, we design the continuous change relation (CCR) branch to further obtain the continuous and detail change relation features to improve the change discrimination capability of the model. The experimental results show that our network performs better, in terms of F1, IoU, and OA, than those of the existing advanced networks for change detection on four public very high-resolution (VHR) remote sensing datasets. 
Our source code is available at {\url{https://github.com/DalongZ/NAME}}.

\end{abstract}

\IEEEpeerreviewmaketitle

\vspace{-0.2cm}
\section{Introduction}
\begin{figure}[t]
	\centering
	\includegraphics[width=0.99\linewidth]{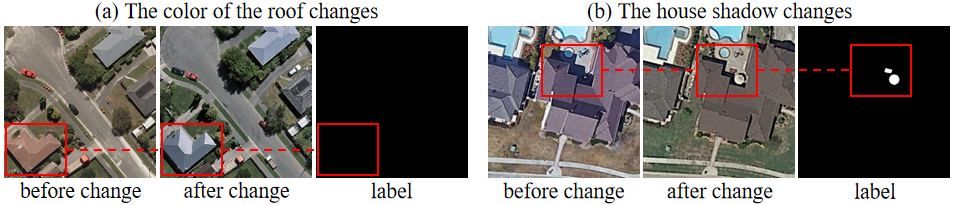}
	\captionsetup{font={small}}
	\caption{A variety of pseudo changes become the challenges in change detection: (a) the roof color changes, and (b) the house shadow changes.}
	\label{fig1}
	\vspace{-0.5cm}
\end{figure}

\vspace{-0.1cm}
\IEEEPARstart{T}{he} definition of change detection is the identification of changes in the surface area found in images over time.
The main problem of change detection is how to extract ``semantic change" and suppress ``non-semantic change" in complex environments. ``Semantic change" mainly refers to the appearance and disappearance of objects which are artificially defined and vary according to specific applications. The causes of ``non-semantic change" often include seasonal changes, changes of illumination, and interferences from object shadows. 
As shown in Fig. \ref{fig1}, ``non-semantic change" caused by illumination and shadows become a challenge in change detection. 
Therefore, many traditional change detection methods cannot extract the effective change features and have the robust detection ability in the complex scenes, which include band difference, change vector analysis \cite{malila1980change}, principal component analysis \cite{zhang2007remote} and slow feature analysis \cite{wu2013slow}.

With the rapid development of deep learning technology, a large number of neural networks and components have been developed for change detection to extract the robust deep features.
Daut et al. \cite{daudt2018fully} provided the three most common baseline networks for change detection. 
Since the single-branch network \cite{daudt2018fully} lacks the detail information of a single image, the architectures for change detection are dominated by the two-branch networks \cite{zhang2020deeply,chen2021remote,li2022transunetcd,feng2022icif,chen2020spatial,9975266}.
At present, the advanced networks usually combine the transformer and CNN in a serial \cite{chen2021remote,li2022transunetcd} or parallel \cite{feng2022icif} manner to obtain  the global information and local information of the image. Some researchers also introduced recurrent neural network \cite{papadomanolaki2021deep} or proposed spatiotemporal attention \cite{chen2020spatial} and video understanding module \cite{9975266} to describe the temporal correlation between two input images.

Although these deep networks have achieved good change detection results, they have some shortcomings. The single-branch architecture concatenates the two input images and then feeds them into the network, which leads to the lacks of detail information and positioning information of a single image \cite{daudt2018fully}. The two-branch architecture often uses the siamese encoder to extract the features of each single image respectively and fuses these features in the decoder stage. 
The architectural contradiction between the two-branch encoder and the single decoder leads to the disappearance of gradient propagation and affects the low-level feature learning of the two original images \cite{zhang2020deeply}.
Moreover, these networks do not explicitly mine the change relationship features, which resulted in the ambiguity of understanding for the change relationship between the two images. 
Motivated by the defects of these models, we propose a triple branch network to respectively mine the three kinds of change features of two input images: pre-changed image features, post-changed image features and change relation features. From the perspective of extracting the global information and local information of the image, we further extract and fuse the three features by combining the transformer and CNN. In the upsampling stage of the decoder, we supplement the detail features from CNN to improve the precise detection performance of the model.

Lin et al. \cite{9975266} constructed the pseudo video frames (PVF) from the input image pairs, and then mined the temporal features for the PVF to enhance the change detection performance of the network. But we do not think that there is some temporal correlation between two images in bi-temporal change detection. For example, the disappearance or appearance of a building belongs to a specific ``semantic change". Deep networks cannot explicitly learn such nonexistent temporal semantic information. However, it is undoubtedly another perspective to abstract the change relation features by constructing and mining the PVF. Based on this inspiration, we design the CCR branch to further provide the model with more diverse change relationship information.

In summary, our main contributions are threefold:
\begin{itemize}
	\item[1)] We firstly propose a triple branch network NAME that explicitly mines the pre-changed image features, post-changed image features and change relation features respectively for change detection. And we extract, interact and fuse these three features by combining the transformer and CNN.
 	\item[2)] The CCR branch is proposed to construct the PVF and learn the continuous and detailed change relationship features by mining the PVF, which further enrichments the change relationship information of the model. It is a plug and play module that has been experimentally proven to be effective not only in our proposed network but also with other change detection networks.
	\item[3)] Through a series of experimental comparisons, our method is superior to other advanced methods on four common public VHR change detection datasets.
\end{itemize}

The remainder of this letter is organized as follows. 
Section \ref{section2} elaborates the proposed NAME network.
%and provides the corresponding optimization algorithm.
The experimental evaluations and ablation studies are given in Section \ref{section3}. 
Section \ref{section4} presents the conclusion of this letter.

\vspace{-0.1cm}
\section{methodology}
\label{section2}
\vspace{-0.1cm}
\subsection{Network Architecture}
\begin{figure*}[t]
	%	\vspace{-0.2cm}
	%	\setlength{\abovecaptionskip}{-0.01cm}  
	%	\setlength{\belowcaptionskip}{-0.2cm}
	\centering
	\includegraphics[width=0.999\linewidth]{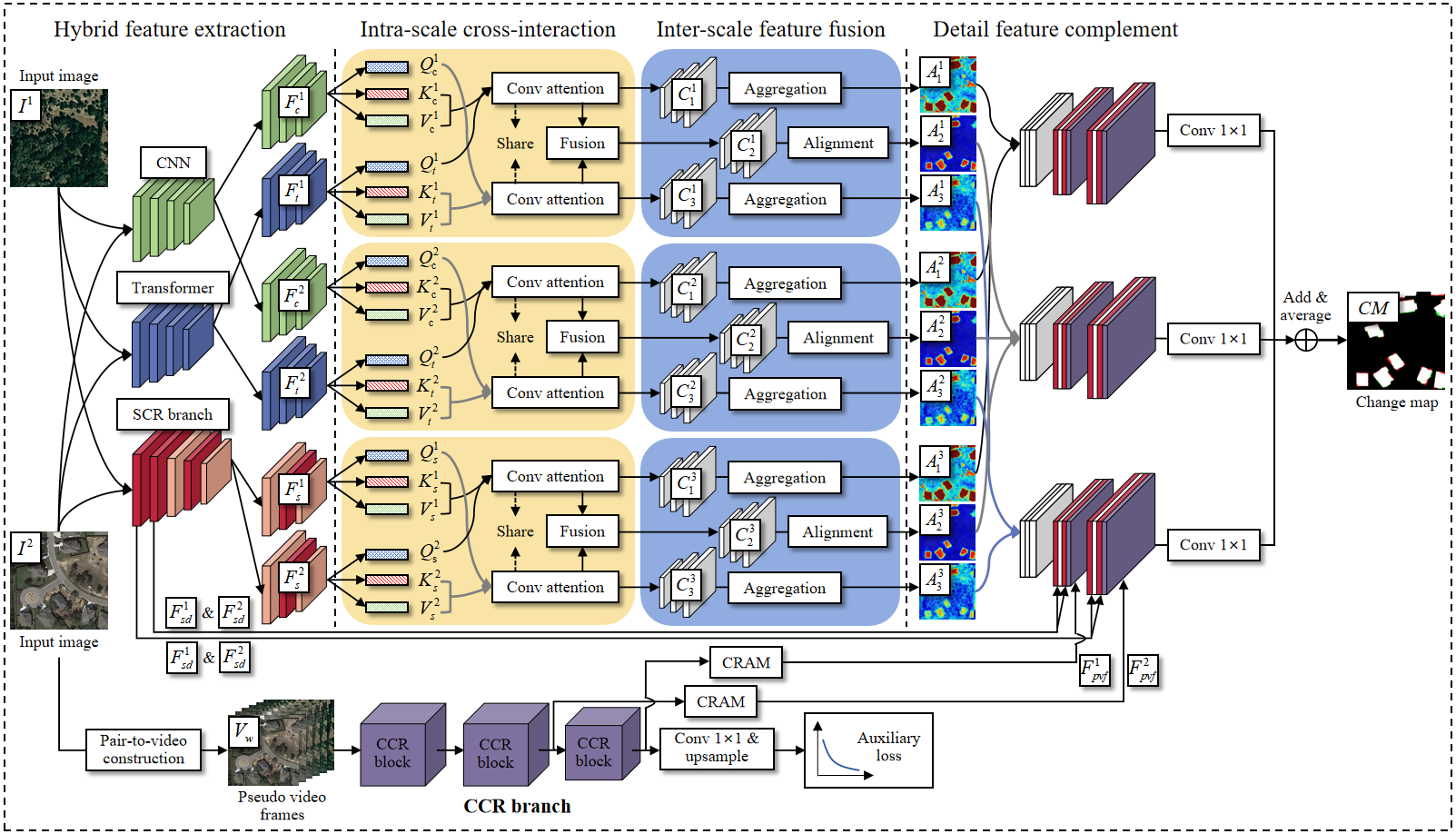}
	\captionsetup{font={small}}
	\caption{The overall architecture of NAME. We propose a triple branch network to extract the pre-changed image features, post-changed image features and structured change relation features of two input images, and design the CCR branch to further mine the continuous change relationship information to improve the detection capability of the model.}
	\label{fig2}
	\vspace{-0.3cm}
\end{figure*}

\vspace{-0.1cm}
The overall architecture of NAME is shown in Fig. \ref{fig2}. 
First of all, we define some characters. ${F}$, ${A}$ and ${D}$ and ${\tilde D}$ denote the features, features of the activation maps and features of the decoder respectively. We use ${\texttt {\bf {CNN}}}$, ${\texttt {\bf {Transformer}}}$, ${\texttt {\bf {SCRB}}}$ and ${\texttt {\bf {CCRB}}}$ to denote CNN branch, the transformer branch, structured change relation (SCR) branch and CCR branch, respectively. ${\texttt {\bf {ISCI}}}$ and ${\texttt {\bf {ISFF}}}$ represent intra-scale cross-interaction (ISCI) and inter-scale feature fusion (ISFF), respectively. ${c}$, ${t}$, ${s}$, ${sd}$ and ${pvt}$ represent the features from CNN branch, the transformer branch, SCR branch, the detail information of SCR branch and CCR branch, respectively.
${i}$ refers to the input image and ${j}$ refers to the output of ISFF.

Then we elaborate on the details of the network. We input two images ${I^1}$ and ${I^2}$ to the different branches respectively for feature extraction: 
\begin{equation}
	\vspace{-0.1cm}
	\label{equation1}
	\left\{ {\begin{array}{*{20}{l}}
			{F_c^1,F_c^2 = \texttt {\bf {CNN}}({I^1},{I^2}),}\\
			{F_t^1,F_t^2 = \texttt {\bf {Transformer}}({I^1},{I^2}),}\\
			{F_s^1,F_s^2,F_{sd}^1,F_{sd}^2 = \texttt {\bf {SCRB}}({I^1},{I^2}).}
	\end{array}} \right.
\end{equation}
Here, ${F_c^1}$ and ${F_c^2}$, ${F_t^1}$ and ${F_t^2}$, ${F_s^1}$ and ${F_s^2}$, and ${F_{sd}^1}$ and ${F_{sd}^2}$ represent the local features, global features, mixed features, and mixed detail features of the images ${I^1}$ and ${I^2}$, respectively. The SCR branch is a mixed serial network composed of the transformer and CNN. The first three pairs of features are fed into the ISCI and ISFF for feature interaction and feature fusion, respectively:
\vspace{-0.1cm}
\begin{equation}
	\vspace{-0.1cm}
	\label{equatio2}
	\left\{ {\begin{array}{*{20}{l}}
			{A_1^i,A_2^i,A_3^i = \texttt {\bf {ISFF}}(\texttt {\bf {ISCI}}(F_c^i,F_t^i)),i = 1,2},\\
			{A_1^3,A_2^3,A_3^3 = \texttt {\bf {ISFF}}(\texttt {\bf {ISCI}}(F_s^1,F_s^2)),}
	\end{array}} \right.
\end{equation}
where nine different features ${A}$ are generated to improve the diversity of model information as described in Fig. \ref{fig2}.

From the perspective of change detection, the features generated by the three branches at this time represent the pre-changed image features, post-changed image features and change relation features respectively. So we concatenate and convolve ${A_j^1}$, ${A_j^2}$ and ${A_j^3}$ to get the feature ${D_j}$:
\vspace{-0.1cm}
\begin{equation}
	\vspace{-0.1cm}
	\label{equatio3}
	{D_j} = \texttt {\bf {Conv}}({\texttt {\bf {Concat}}}(A_j^1,A_j^2,A_j^3)),j = 1,2,3.
\end{equation}
We upsample ${D_j}$, and obtain ${F_{pvt}}$ by ${F_{pvt} = \texttt {\bf {CCRB}}({I^1},{I^2})}$. 
They are concatenated with ${F_{sd}^1}$ and ${F_{sd}^2}$, and the convolution operation is then performed:
\vspace{-0.1cm}
\begin{equation}
	\vspace{-0.1cm}
	\label{equatio4}
	{{\tilde D}_j} = \texttt {\bf {Conv}}(\texttt {\bf {Concat}}(F_{sd}^1,\texttt {\bf {Up}}({D_j}),F_{sd}^2,{F_{pvt}})).
\end{equation}
Here, Eq. (\ref{equatio4}) needs to be executed twice. Finally, we get the probability map of each branch by convolution ${1\times1}$ and sigmoid function, and then obtain the change map ${CM}$ by summing and averaging the three probability maps:
\vspace{-0.1cm}
\begin{equation}
	\vspace{-0.1cm}
	\label{equatio5}
	CM = (\sum\limits_{j = 1}^3 {\texttt {\bf {Sigmoid}}(\texttt {\bf {Conv}}{\bf 1\times1}({{\tilde D}_j})))/3}. 
\end{equation}

In summary, we design a parallel branch of the transformer and CNN to extract the pre-changed image features and post-changed image features, and a serial branch of the transformer and CNN to mine the change relation features. In the decoder stage, these three kinds of features are concatenated and supplemented with the detail features ${F_{sd}^1}$, ${F_{sd}^2}$ and ${F_{pvt}}$ from the serial branch and CCR branch. Finally, ensemble learning is used to generate the change map. 
At the same time, the extraction, interaction and fusion of global information from the transformer and local information from CNN of the images make the detection performance of the model advanced and robust.
The ISCI and ISFF are presented in Appendix {\textcolor[rgb]{1 0 0}{A}}.

\vspace{-0.2cm}
\subsection{Structured Change Relation Branch}
\vspace{-0.1cm}
\begin{figure}[t]
	%		\vspace{-0.2cm}
	%\setlength{\abovecaptionskip}{-0.01cm}  
	\setlength{\belowcaptionskip}{-0.1cm}
	\centering
	\includegraphics[width=0.9\linewidth]{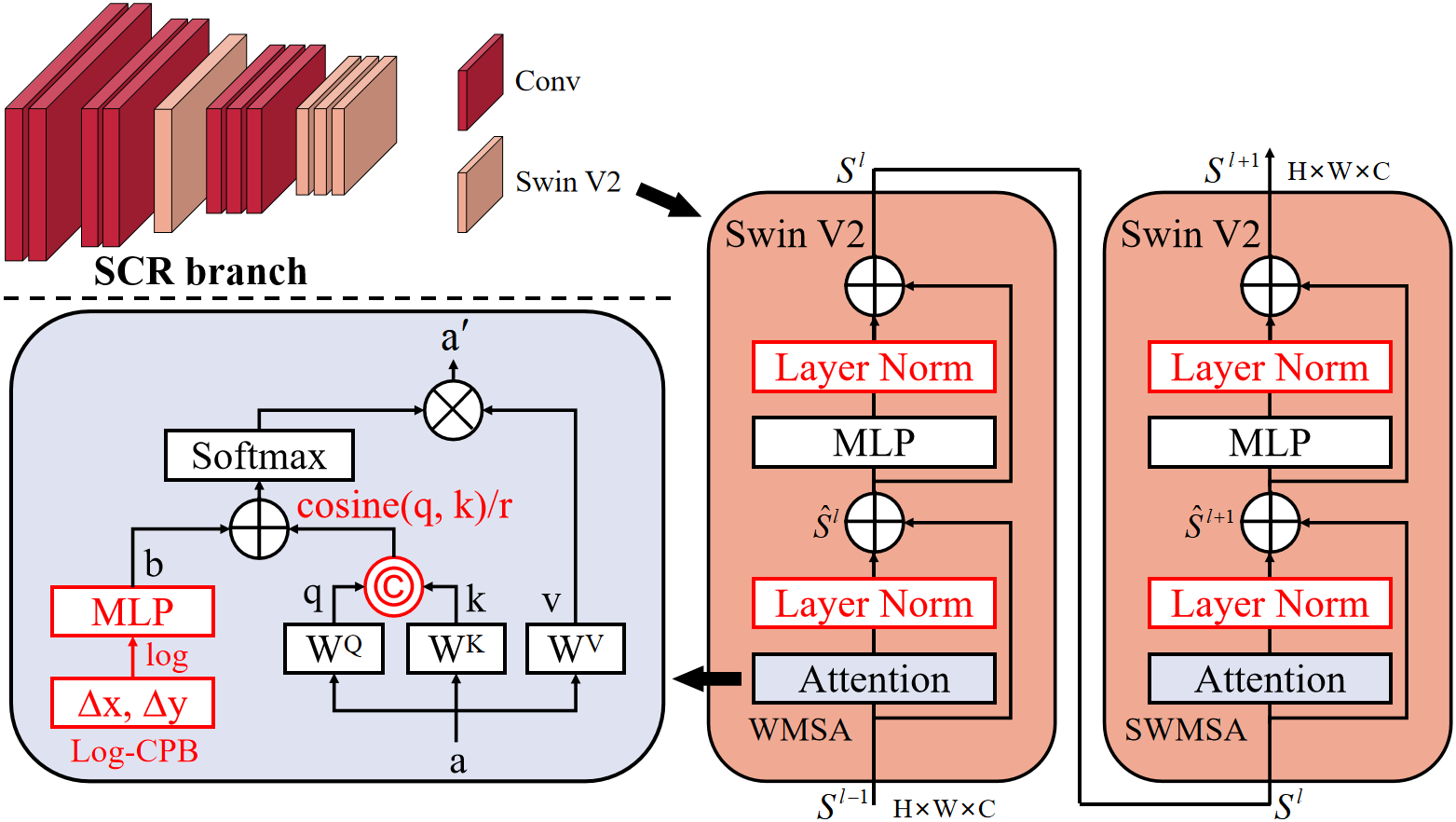}
	\captionsetup{font={small}}
	\caption{{We design the SCR branch by combining the Swin-V2 blocks and convolution blocks.}}
	\label{fig3}
	\vspace{-0.3cm}
\end{figure}

We use pyramid vision transformer V2-B1 \cite{wang2022pvt} as the backbone of the transformer branch and ResNet18 \cite{he2016deep} as the backbone of the CNN branch. 
In order to enhance the features diversity in the whole model, we design the SCR branch by combining the swin transformer V2 (Swin-V2) blocks and convolution blocks. 
As shown in Fig. \ref{fig3}, the SCR branch consists of five encoder stages, where the encoder of stage 1, 2 and 4 contains 2, 2 and 3 convolution blocks, while stage 3 and 5 contains 1 and 3 Swin-V2 blocks.
We construct the serial hybrid network combining the transformer and CNN to extract the deep features of the two input images, which contain both global information and local information of the images. Then, these two deep features are interacted and fused to mine the structured change relationship information.

Swin-type transformers reduce the number of model parameters while modeling global information through shifted window and hierarchical mechanism \cite{liu2021swin}. Swin-V2 \cite{liu2022swin} further employs the post-normalization and scaled cosine attention techniques to improve the stability of the large vision model. 
At the same time, the log-spaced continuous position bias method is used to alleviate the problem of transferring the model trained on low-resolution images to high-resolution images. So we use Swin-V2 as the base block to build the SCR branch. The Swin-V2 is described in detail in Fig. \ref{fig3} and Appendix {\textcolor[rgb]{1 0 0}{B}}.

\vspace{-0.2cm}
\subsection{Continuous Change Relation Branch}
\vspace{-0.1cm}
We do not agree that deep learning can extract the nonexistent ``temporal semantic change" of two images for change detection \cite{9975266}, but it is undoubtedly a supplement to the complete change features that mining the continuous change relationship information by constructing the PVF from two images. Different from the temporal encoder in \cite{9975266}, the CCR branch focuses more on the detail change relationship features to better extract the rich ground object information for the current VHR remote sensing images.
Secondly, we design the change relationship aggregation module (CRAM) to fuse the continuous change relationship information before introducing it into the triple branch network. 
The ablation experiment in Table \ref{tab3} proves that the features extracted by the temporal encoder do not have a performance gain for the whole model, while our proposed CCR branch improves the detection ability of the model.
Because the abstract features of size 32$\times$32 and 64$\times$64 from the temporal encoder have been covered by the information of the triple branch network, while the detail features of size 128$\times$128 and 256$\times$256 extracted by the CCR branch contain more abundant ground object information in the VHR images.
\begin{table*}[h!]
	\captionsetup{font={small}}
	\caption{\centering{The comparison results on the four change detection datasets. The best values are highlighted in \newline bold font. All results are expressed as percentages ($\%$).}}
	\vspace{-0.2cm}
	\centering
	\scalebox{0.67}{
		\renewcommand\arraystretch{1}
		%		\hspace{-2mm}
		\setlength{\tabcolsep}{3mm}{
			\begin{tabular}{c|cccc}
				\hline
				\multirow{2}{*}{\textbf{Method}} & \textbf{LEVIR-CD}                     & \textbf{SVCD}                & \textbf{WHU-CD}        & \textbf{SYSU-CD}  \\
				& Pre. / Rec. / F1 / IoU / OA    & Pre. / Rec. / F1 / IoU / OA    & Pre. / Rec. / F1 / IoU / OA & Pre. / Rec. / F1 / IoU / OA  \\ \hline
				FC-EF                            & 86.16 / 86.20 / 86.18 / 76.16 / 98.59 & 85.35 / 77.56 / 81.27 / 42.14 / 95.59     & 86.13 / 86.01 / 86.07 / 75.67 / 98.82& 79.30 / 68.84 / 73.70 / 44.64 / 88.41   \\
				FC-Siam-Diff                     & 90.36 / 84.81 / 87.50 / 81.06 / 98.77 & 92.28 / 78.70 / 84.95 / 48.87 / 96.56     & 81.40 / 89.11 / 85.08 / 71.79 / 98.68 &\textbf{89.80} / 58.49 / 70.84 / 42.37 / 88.64 \\
				FC-Siam-Conc                     & 87.30 / 87.81 / 87.55 / 75.09 / 98.73 & 92.04 / 81.94 / 86.70 / 52.95 / 96.90     & 79.98 / 90.94 / 85.11 / 66.25 / 98.65&76.19 / 77.98 / 77.07 / 50.39 / 89.06  \\ \hline
				IFNet                            & \textbf{93.73} / 87.31 / 90.40 / \textbf{84.04} / 99.06 & 97.71 / 93.64 / 95.63 / 81.31 / 98.94     & \textbf{98.51} / 82.46 / 89.77 / \textbf{90.28} / 99.21& 85.16 / 75.36 / 79.96 / 57.22 / 91.09   \\
				SNUNet-CD                        & 91.00 / 88.30 / 89.63 / 79.18 / 98.96 & 98.13 / 97.62 / 97.87 / 88.24 / 99.48     & 88.50 / 90.31 / 89.40 / 73.50 / 99.09 & 80.03 / 76.62 / 78.29 / 52.99 / 89.98  \\
				BIT                              & 91.81 / 88.00 / 89.86 / 79.35 / 98.99 & 97.07 / 96.43 / 96.75 / 83.52 / 99.20     & 92.10 / 92.41 / 92.26 / 78.56 / 99.34& 81.67 / 76.52 / 79.01 / 52.41 / 90.41  \\
				TransUNetCD                      & 90.62 / 88.44 / 89.52 / 79.63 / 98.94 & 97.44 / 96.52 / 96.98 / 84.81 / 99.26     & 94.72 / 91.21 / 92.93 / 83.82 / 99.41& 77.25 / 80.17 / 78.68 / 54.72 / 89.75  \\
				ICIF-Net                         & 91.85 / 90.01 / 90.92 / 83.37 / 99.08 & 98.09 / 97.92 / 98.00 / 88.87 / 99.51     & 94.83 / 92.74 / 93.77 / 86.42 / 99.48 & 83.36 / 76.01 / 79.52 / 55.29 / 90.77  \\
				FCCDN                            & 92.10 / 84.86 / 88.33 / 80.48 / 98.86 & 95.96 / 95.56 / 95.76 / 79.79 / 98.96     & 92.65 / 90.44 / 91.53 / 79.94 / 99.29 &78.57 / 78.14 / 78.36 / 53.33 / 89.82  \\ \hline
				Ours                             & 92.29 / \textbf{90.87} / \textbf{91.58} / 83.50 / \textbf{99.15} & \textbf{98.46} / \textbf{98.28} / \textbf{98.37} / \textbf{90.74} / \textbf{99.60}     & 96.25 / \textbf{92.76} / \textbf{94.47} / 87.36 / \textbf{99.54} &81.31 / \textbf{80.80} / \textbf{81.05} / \textbf{58.40} / \textbf{91.09} \\ \hline
		\end{tabular}}
	}
	\label{tab1}
	\vspace{-0.3cm}
\end{table*}

We construct the pseudo video frames by linear interpolation:
\vspace{-0.2cm}
\begin{equation}
	\vspace{-0.1cm}
	\label{equatio6}
	{V_w} = \varphi ({I^1},{I^2},w) = {I^1} + \frac{w}{{W - 1}}({I^2} - {I^1}), 
\end{equation}
where the constructed pseudo video comprises $W$ frames in total, and the $w$-th ($0 \le w<W$) frame $V_w$ is deduced using the images $I^1$ and $I^2$, along with the frame index $w$. $\varphi$ is the constructed function of the PVF.
The PVF represents continuous change relation information. We then extract the continuous change relation features through three CCR blocks. The supervision loss is imposed on the output of the third CCR block to guide the optimization of the CCR branch. 
Finally, the feature maps of the second and third CCR blocks are transformed by the CRAM and then forwarded to the last two layers of the decoder in the triple branch network, respectively. The dimensions of $F_{{\rm{pvf}}}^1$ and $F_{{\rm{pvf}}}^2$ are 128$\times$128$\times$64 and 256$\times$256$\times$64, respectively. Since the triple branch network adopts the idea of ensemble learning, $F_{{\rm{pvf}}}^1$ and $F_{{\rm{pvf}}}^2$ need to be added to three different decoders at the same time.
The CCR block and CRAM are elaborated in Appendix {\textcolor[rgb]{1 0 0}{C}}.

\vspace{-0.3cm}
\subsection{Loss Function}
\vspace{-0.1cm}
\label{section3_4}
Bi-temporal change detection is fundamentally a binary classification task, so binary cross entropy (BCE) loss is usually used as in the following:
\vspace{-0.2cm}
\begin{equation}
	\vspace{-0.15cm}
	\label{equation22}
	{L_{BCE}} =  - (n\log (\hat n) + (1 - n)\log (1 - \hat n)),
\end{equation}
where ${n}$ and ${\hat n}$ denote the predicted change confidence and the label in the corresponding position, respectively. To mitigate the class imbalance problem in change detection, Dice loss is often used:
\vspace{-0.2cm}
\begin{equation}
	\vspace{-0.15cm}
	\label{equation23}
	{L_{Dice}} = 1 - \frac{{2\hat nn + \sigma }}{{\hat n + n + \sigma }}.
\end{equation}
Here, adding ${\sigma}$ avoids the case where the denominator is zero, and ${n}$ and ${\hat n}$ are similarly defined as in Eq. (\ref{equation22}).
The loss function used in our model is a combination of BCE and Dice loss. Moreover, the total loss function is expressed as follows:
\vspace{-0.3cm}
\begin{equation}
	\vspace{-0.15cm}
	\label{equation24}
	{L_{Total}} = (\sum\limits_{m = 1}^4 {L_{BCE}^m}  + {\lambda}L_{Dice}^m)/4,
\end{equation}
where $m$ represents the three branches and CCR branch in the network, and ${\lambda}$ is the weight coefficient and is set to 0.5.

\vspace{-0.2cm}
\section{Experimental Evaluation}
\label{section3}
\vspace{-0.1cm}
\subsection{Experimental Configurations}
\vspace{-0.1cm}
To fully verify the performance of our model, we conducted a large number of comparative experiments and ablation studies on four public commonly used VHR datasets. The datasets include: learning, vision, and remote sensing change detection dataset (LEVIR-CD), season-varying change detection dataset (SVCD), wuhan university change detection dataset (WHU-CD), and sun yat-sen university change detection dataset (SYSU-CD). 
The images of these datasets have a very high spatial resolution, ranging from 0.03 to 1 meter per pixel. 
We performed the data augmentation on the SVCD using rotation and flip operations. The images of all datasets are cropped into 256 × 256 image patches. 

The comparison methods include FC-EF, FC-Siam-Diff, FC-Siam-Conc  \cite{daudt2018fully}, IFNet \cite{zhang2020deeply}, SNUNet-CD \cite{fang2021snunet}, BIT \cite{chen2021remote}, TransUNetCD \cite{li2022transunetcd}, ICIF-Net \cite{feng2022icif} and FCCDN \cite{chen2022fccdn}. The first three methods are commonly used baselines and the last six are recent deep learning algorithms in change detection.
In terms of quantitative index evaluation, we mainly focused on F1 score and employed precision, recall, IoU and OA.
We implemented the NAME using Pytorch framework and conducted the experiments on a single NVIDIA GeForce RTX 3090 GPU. See our publicly available source code for more specific hyperparameters.
\vspace{-0.3cm}
\subsection{Comparison and Analysis of Experiment Results}
\begin{figure*}[t]
%	\vspace{-0.3cm}
	\setlength{\abovecaptionskip}{-0.01cm}  
	\setlength{\belowcaptionskip}{-0.5cm}
	\centering
	\includegraphics[width=0.99\linewidth]{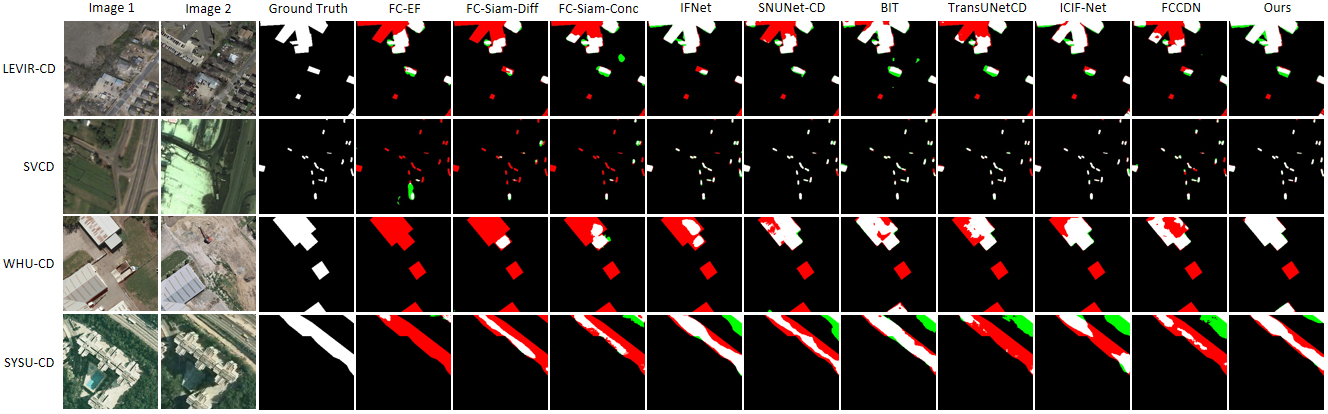}
	\captionsetup{font={small}}
	\caption{The visualization results of various methods on the LEVIR-CD, SVCD, WHU-CD and SYSU-CD test sets. We use different colors to represent TP(white), TN(black), FP(green), and FN(red) in the change maps.}
	\label{fig4}
\end{figure*}

\vspace{-0.1cm}
The analysis of the experimental results in Table \ref{tab1} shows that our model is more advanced than other methods on four mainstream change detection datasets. 
In terms of F1 score, NAME leads the method in second place by 0.66(LEVIR-CD), 0.37(SVCD), 0.70(WHU-CD), and 1.09(SYSU-CD), respectively. 
Although FC-Siam-Diff and IFNet pay more attention to the changed regions in some scenes and have occasional advantages in the precision and IoU, our model undoubtedly shows the more accurate detection capability and robustness to deal with various scenes in the comprehensive indicators F1 and OA. As shown in Fig. \ref{fig4}, our model performs well for both structured and small objects. 
Especially compared with the three advanced deep networks combining the transformer and CNN, BIT, TransUNetCD and ICIF-Net, the explicit mining of change relation information and the emphasis on detail information significantly improve the change discrimination ability of our model.
\vspace{-0.3cm}
\subsection{Ablation Study}
\vspace{-0.1cm}
We use F1 score as the main metric in ablation experiments. OA is positively correlated with F1 score, while IoU presents some uncertainty. So we bold all F1 scores in the tables.
\subsubsection{Effectiveness of Overall Network}
Through the ablation study of the overall network in Table \ref{tab2}, it is obvious that the use of the SCR branch or CCR branch has a significant improvement compared to the baseline. Although there is a phenomenon of mutual cancellation when they work together, their cooperation still improves the overall performance of the model.
\begin{table}[h!]
	\captionsetup{font={small}}
	\caption{\centering{The ablation study for the overall network on the two datasets. All results are expressed as percentages ($\%$).}}
	\vspace{-0.2cm}
	\centering
	\scalebox{0.75}{
		\renewcommand\arraystretch{1}
		\begin{tabular}{cccc}
			\hline
			\multicolumn{2}{c}{\makebox[0.16\textwidth][c]{\textbf{Overall Network}}}                  
			& \textbf{LEVIR-CD} 
			& \textbf{SYSU-CD}\\ 		 
			\makebox[0.09\textwidth][c]{SCR Branch}  & \makebox[0.09\textwidth][c]{CCR Branch}                           
			& F1 / IoU / OA                                    
			& F1 / IoU / OA                                 \\ \hline
			&                               
			&  \textbf{90.92} / 83.37 / 99.08        
			&  \textbf{79.52} / 55.29 / 90.77  \\ 
			\checkmark           & \multicolumn{1}{c}{}  
			& \textbf{91.42} / 83.69 / 99.14     
			&   \textbf{80.90} / 57.34 / 91.17 \\	
			& \checkmark            
			& \textbf{91.35} / 82.57 / 99.13          
			& \textbf{80.87} / 58.14 / 91.02   \\	
			\checkmark &  \checkmark                 
			& \textbf{91.58} / 83.50 / 99.15      
			& \textbf{81.05} / 58.40 / 91.09  \\ \hline		
		\end{tabular}
	}
	\label{tab2}
	\vspace{-0.1cm}
\end{table}
\begin{table}[h!]
	\captionsetup{font={small}}
	\caption{\centering{The ablation study for the CCR branch on the two datasets. All results are expressed as percentages ($\%$).}}
	\vspace{-0.2cm}
	\centering
	\scalebox{0.75}{
		\renewcommand\arraystretch{1}
		\begin{tabular}{cccc}
			\hline
			%\multicolumn{2}{c}{\makebox[0.16\textwidth][c]{\textbf{CCR Branch}}} 
			 \multirow{2}{*}{\textbf{Branch}}     &\multirow{2}{*}{\textbf{Frames}}&  \textbf{LEVIR-CD} & \textbf{SYSU-CD}\\ 		               
			&    & F1 / IoU / OA      & F1 / IoU / OA    \\ \hline
			None      &  & \textbf{91.42} / 83.69 / 99.14     &   \textbf{80.90} / 57.34 / 91.17 \\			
			Temporal Branch     &4	& \textbf{91.40} / 85.09 / 99.14     &  \textbf{80.90} / 57.41 / 91.17 \\			
			CCR Branch   &2	& \textbf{91.51} / 83.64 / 99.15    	&  \textbf{80.99} / 57.52 / 91.23    \\
			CCR Branch   &3	& \textbf{91.53} / 83.22 / 99.14    	&  \textbf{81.27} / 58.25 / 91.25   \\
			CCR Branch   &4 & \textbf{91.58} / 83.50 / 99.15    	& \textbf{81.05} / 58.40 / 91.09    \\ \hline	
		\end{tabular}
	}
	\label{tab3}
	\vspace{-0.2cm}
\end{table}

%\vspace{-0.2cm}
\subsubsection{Impact of CCR Branch}
The ablation study in Table \ref{tab3} shows that temporal branch does not bring the performance gain for our model, since the abstract features of the triple branch network already cover the features from temporal branch. However, the role of CCR branch is obviously important. As the number of frames increases, the positive impact of CCR branch tends to improve. See Appendix {\textcolor[rgb]{1 0 0}{D}} for the performances of other models combined with CCR branch.

$3)$  The computational complexity, training process and network visualization of our model, the ablation study of SCR branch, and the supplement of the change maps of various methods on the four datasets in Appendix {\textcolor[rgb]{1 0 0}{E}} to {\textcolor[rgb]{1 0 0}{I}}.

\vspace{-0.2cm}
\section{Conclusions}
\vspace{-0.1cm}
\label{section4}
In this paper, we propose a concept of explicitly mining change relation information for change detection. Based on this concept, we design the SCR branch and CCR branch to extract the structured and continuous change relationship features respectively, which significantly improves the detection capability of the whole model. Meanwhile, the emphasis on detail features is also reflected in our architecture. 
The experiments verify that the CCR branch can be seamlessly integrated with different change detection networks to extract the continuous change relationship information to enhance network performance.
We hope that our work bring new heuristics for mining change relationships in change detection.

\vspace{-0.2cm}
{\footnotesize
\bibliographystyle{IEEEtran}
\bibliography{introduction}}
\end{document}